\documentclass{article}

\PassOptionsToPackage{numbers, sort&compress}{natbib}

\usepackage[preprint]{neurips_2021}

\usepackage[utf8]{inputenc} 
\usepackage[T1]{fontenc}    
\usepackage{hyperref}       
\usepackage{url}            
\usepackage{booktabs}       
\usepackage{amsfonts}       
\usepackage{nicefrac}       
\usepackage{microtype}      
\usepackage{xcolor}         

\usepackage[pdftex]{graphicx}
\usepackage{bm}
\usepackage{subfigure}
\usepackage{amsmath}
\usepackage{algorithm}
\usepackage{algorithmic}
\usepackage{natbib}
\usepackage{threeparttable}
\usepackage{hyperref}

\title{AutoAdversary: A Pixel Pruning Method for Sparse Adversarial Attack}

\author{
Jinqiao Li\\
National Key Laboratory for Novel Software Technology, Nanjing University, China\\
\texttt{lijq@smail.nju.edu.cn}\\
\And
Xiaotao Liu\\
National Key Laboratory for Novel Software Technology, Nanjing University, China\\
\texttt{liuxiaotao863@gmail.com}\\
\And
Jian Zhao\\
School of Electronic Science and Engineering, Nanjing University, China\\
\texttt{jianzhao@nju.edu.cn}\\
\And
Furao Shen\thanks{Contact Author}\\
National Key Laboratory for Novel Software Technology, Nanjing University, China\\
\texttt{frshen@nju.edu.cn}\\
}

\begin{document}

\maketitle

\begin{abstract}
Deep neural networks (DNNs) have been proven to be vulnerable to adversarial examples. A special branch of adversarial examples, namely sparse adversarial examples, can fool the target DNNs by perturbing only a few pixels. However, many existing sparse adversarial attacks use heuristic methods to select the pixels to be perturbed, and regard the pixel selection and the adversarial attack as two separate steps. From the perspective of neural network pruning, we propose a novel end-to-end sparse adversarial attack method, namely AutoAdversary, which can find the most important pixels automatically by integrating the pixel selection into the adversarial attack. Specifically, our method utilizes a trainable neural network to generate a binary mask for the pixel selection. After jointly optimizing the adversarial perturbation and the neural network, only the pixels corresponding to the value 1 in the mask are perturbed. Experiments demonstrate the superiority of our proposed method over several state-of-the-art methods. Furthermore, since AutoAdversary does not require a heuristic pixel selection process, it does not slow down excessively as other methods when the image size increases.
\end{abstract}


\section{Introduction}


Despite the success of deep neural networks (DNNs), they still face many challenges such as the threat of adversarial examples~\citep{759851e20d2e47aaad2a560211f6a126,goodfellow2014explaining}. An adversarial example is a carefully crafted image to fool the target model by adding small adversarial perturbations to the original clean image. 

In most cases, the $l_p$-norm of the adversarial perturbation is constrained to make it imperceptible to the human eye. Many adversarial attacks~\citep{carlini2017towards,madry2018towards} perturb all pixels of the image under $l_{\infty}$-norm or $l_2$-norm constrain, which is called dense attack. Interestingly, some recent works~\citep{papernot2016limitations,su2019one,dong2020greedyfool} observed that the DNN model can also be fooled if only partial pixels\footnote{In this paper, three channels of one position in an RGB image are regarded as three independent units, each unit is called a pixel.} are perturbed, dubbed sparse attacks. 
Sparse adversarial attacks can only perturb the pixels that have the greatest impact on the classification result, so it can often reduce the perturbation without reducing the attack effect.
Understanding the vulnerability of DNNs under such simple sparse perturbations can further help design methods to improve the robustness of DNN models.

\begin{figure}
	\centering
	\includegraphics[width=0.65\linewidth]{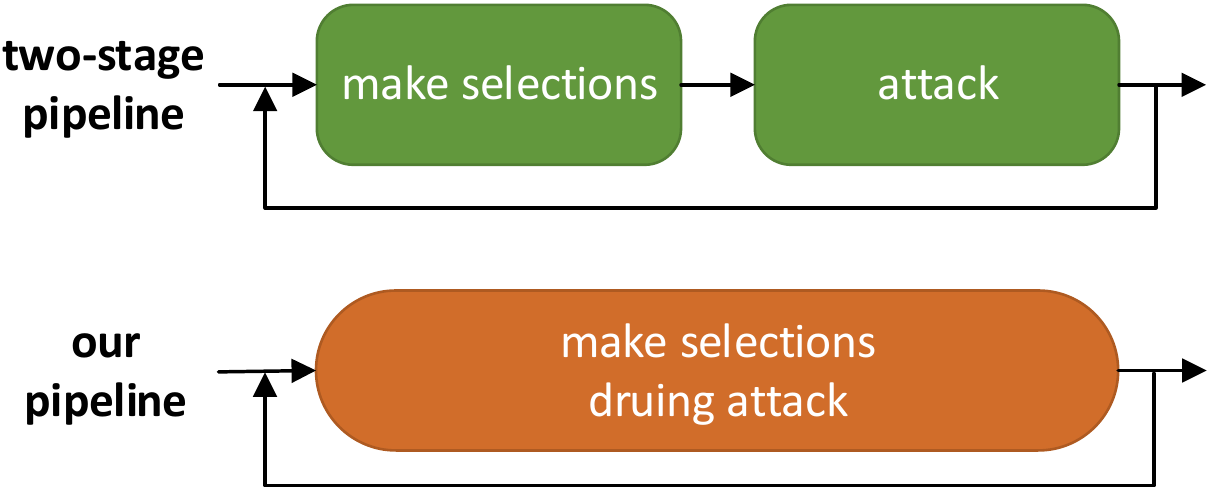}
	\caption{Overview of adversarial attack under $l_0$-norm constraints. The first row is a typical two-stage pipeline, which regards pixels selection and attack as two separate steps. In our method, we integrate pixels selection into attack. During attack, our method automatically removes those pixels that are not important for constructing the adversarial examples.}
	\label{fig_stage}
\end{figure}

Sparse adversarial perturbations are constrained by $l_0$-norm, which makes sparse attack an NP-hard problem. 
As illustrated in the first row of Figure~\ref{fig_stage}, most of the existing methods can be described by a two-stage pipeline. First, these methods use a hand-crafted evaluation criterion to measure the importance of pixels. 
The importance is generally determined by factors such as gradients and perturbation magnitudes. 
According to the importance scores, these methods use heuristic strategies to select pixels. Then these methods re-attack in the new modifiable pixels. These steps will be repeated until the attack is completed. In this type of method, hand-crafted importance evaluation criteria play a vital role. However, hand-crafted criteria may not be suitable for all situations. Moreover, pixels selection and attack are two separate steps in this pipeline. So an interesting question arises: can we use attack to guide pixel selection directly? In other words, can we use an automatic method to select the pixels that need to be perturbed without relying on hand-crafted rules?

Researchers in the neural network pruning task answered a similar question. At the beginning, most neural network pruning methods separate the filters selection and model fine-tuning, and try to find better evaluation criteria for measuring the importance of filters~\citep{liu2017learning,yu2018nisp}. Luo et al.~\citep{luo2020autopruner} proposed an end-to-end and trainable neural network pruning framework called AutoPruner. This framework uses fine-tuning to guide the selection of filters, eliminating the need for hand-crafted importance criteria.

Inspired by AutoPruner, we propose a novel end-to-end sparse adversarial attack method called AutoAdversary. By integrating the pixel selection into attack, the pixels that are not important for constructing adversarial examples can be automatically filtered without any hand-craft rules. Specifically, our method utilizes a trainable neural network whose input is the adversarial perturbation to generate a binary mask for the pixel selection. After joint optimization the adversarial perturbation and the neural network, only the pixels corresponding to the value 1 in the mask are perturbed. Hence, a sparse adversarial perturbation is generated.

Experimental results on CIFAR-10~\citep{krizhevsky2009learning} and ImageNet~\citep{deng2009imagenet} demonstrate that AutoAdversary outperforms previous state-of-the-art approaches. 
In summary, the main contributions of this paper are as follows:
\begin{itemize}
\item \textbf{Innovatively use neural network pruning methods for sparse adversarial attack.} 
According to the inherent similarity between neural network pruning and sparse adversarial attack, we propose an end-to-end sparse adversarial attack method which utilize attack to guide the automatic selection of pixels without hand-crafted rules.
\item \textbf{Under the double constraints of $l_0$-norm and $l_{\infty}$-norm, our method still has a good effect.} 
Our method can not only make the $l_0$-norm of the perturbations as small as possible, but also make the $l_{\infty}$-norm of the perturbations within specified values to avoid being perceived.
\item \textbf{Attacks on large images are still fast.} 
Compared with other methods, our method does not use heuristic strategies to select pixels and requires fewer iterations. So the speed of our method is competitive, and does not slow down excessively as the image size increases.
\end{itemize}

\section{Related work}

\label{section:Related work}
The main difficulty of sparse adversarial attack is how to determine the pixels that need to be perturbed. Most of the existing approaches can be described with a two-stage pipeline. For example, based on the saliency map, JSMA~\citep{papernot2016limitations} uses a heuristic strategy to iteratively select the pixels to be perturbed. 
In each iteration, C$\&$W-$l_0$~\citep{carlini2017towards} first uses the attack under the $l_2$-norm constraint, and then fixes a few least important pixels based on the perturbation magnitudes and gradients. PGD-$l_0+l_{\infty}$~\citep{croce2019sparse} project the perturbations generated by PGD~\citep{madry2018towards} to the $l_0$-ball to achieve the $l_0$ version PGD. The specific projection method is to fix some pixels that do not need to be perturbed according to the perturbation magnitudes and projection loss. SparseFool~\citep{modas2019sparsefool} convert the problem into an $l_1$-norm constraint problem and selects some pixels to perturb in each iteration according to the geometric relationship. Based on the gradient and the distortion map
, GreedyFool~\citep{dong2020greedyfool} selects some pixels to be added to the modifiable pixel set in each iteration, and then uses the greedy method to drop as many less important pixels as possible to obtain better sparsity.

All these methods are trying to find a better way to evaluate pixel importance. However, they both regard pixels selection and attack as two independent steps. We argue that combining these two steps is a better choice.


There are also some methods that beyond the two-stage pipeline. One-Pixel attack~\citep{su2019one} uses the differential evolution algorithm to explored the extreme case that only one pixel is perturbed. But the attack success rate of this method is low. SAPF~\citep{fan2020sparse} is similar to our method. This method factorizes the perturbation into the product of the perturbation magnitude and the binary mask, and uses $l_p$-box ADMM~\citep{wu2018ell} to jointly optimize them. Our method also uses the binary mask, but we generate the mask through a trainable neural network. 
Experimental results show that our method has better sparsity, and the speed on CIFAR-10 is 25 times faster than SAPF.


\section{Sparse adversarial attack}

\begin{figure*}
	\centering 
	\subfigure[original]{
		\label{fig_epsilon_ori}
		\includegraphics[width=0.18\linewidth]{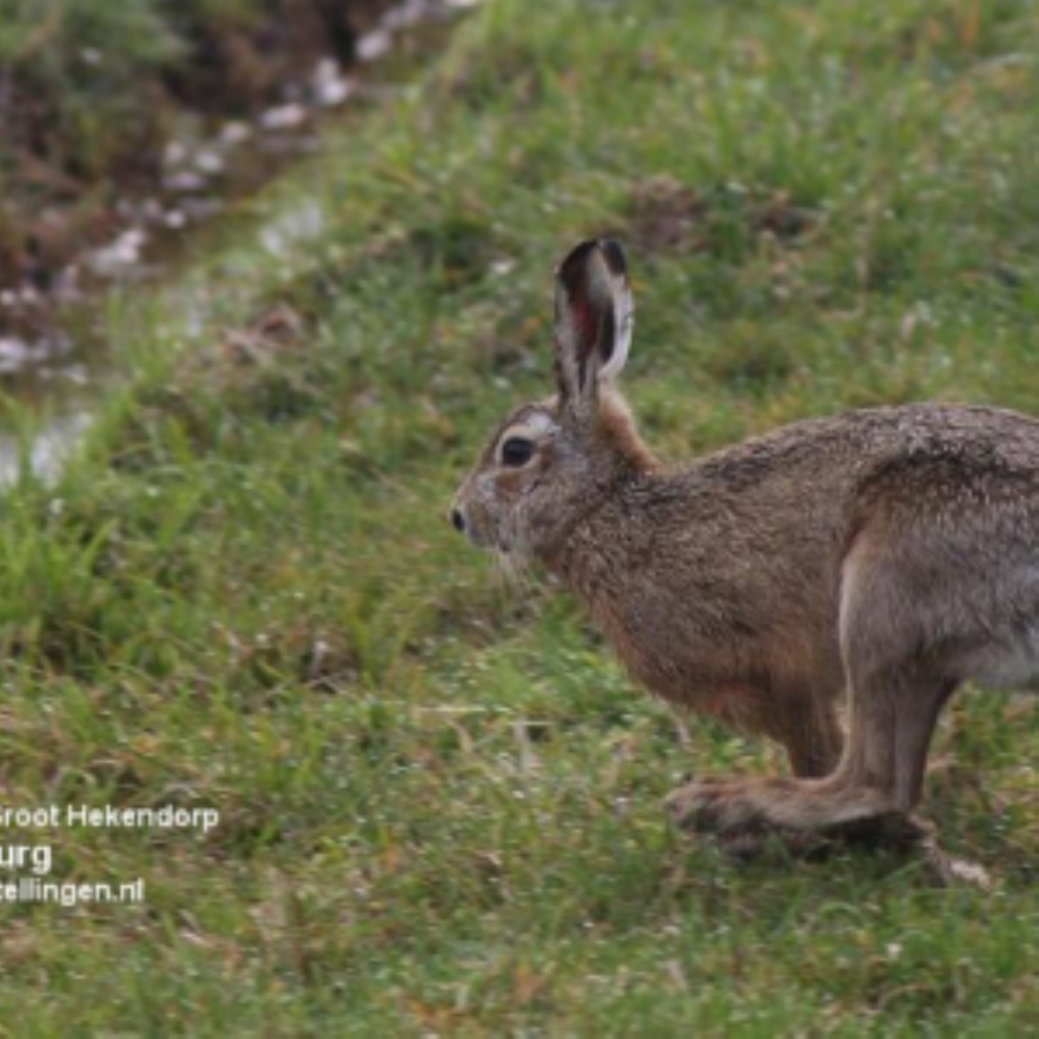}}
	\subfigure[$\epsilon=128/255$]{
		\label{fig_epsilon_128}
		\includegraphics[width=0.36\linewidth]{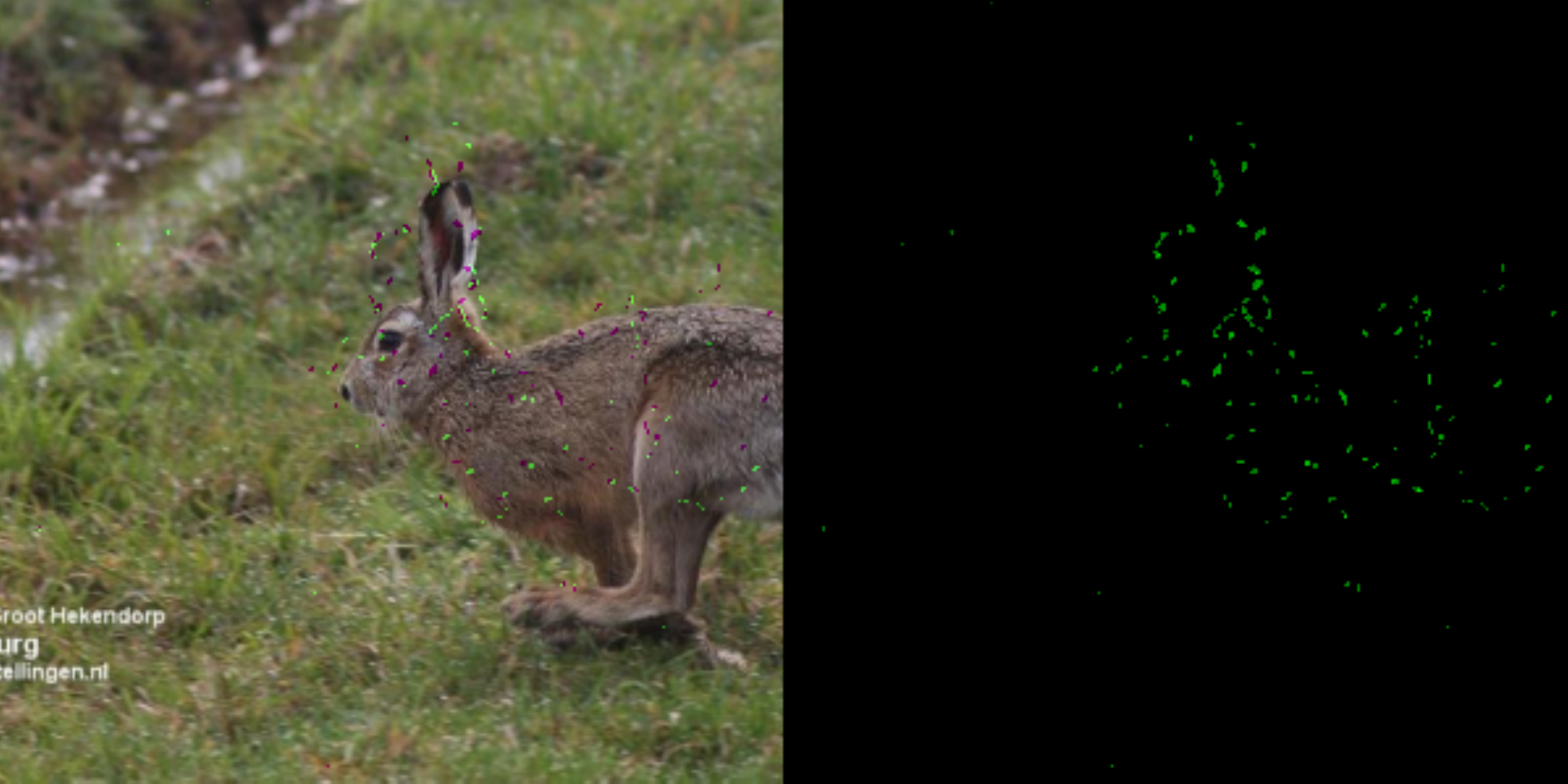}}
	\subfigure[$\epsilon=4/255$]{
		\label{fig_epsilon_4}
		\includegraphics[width=0.36\linewidth]{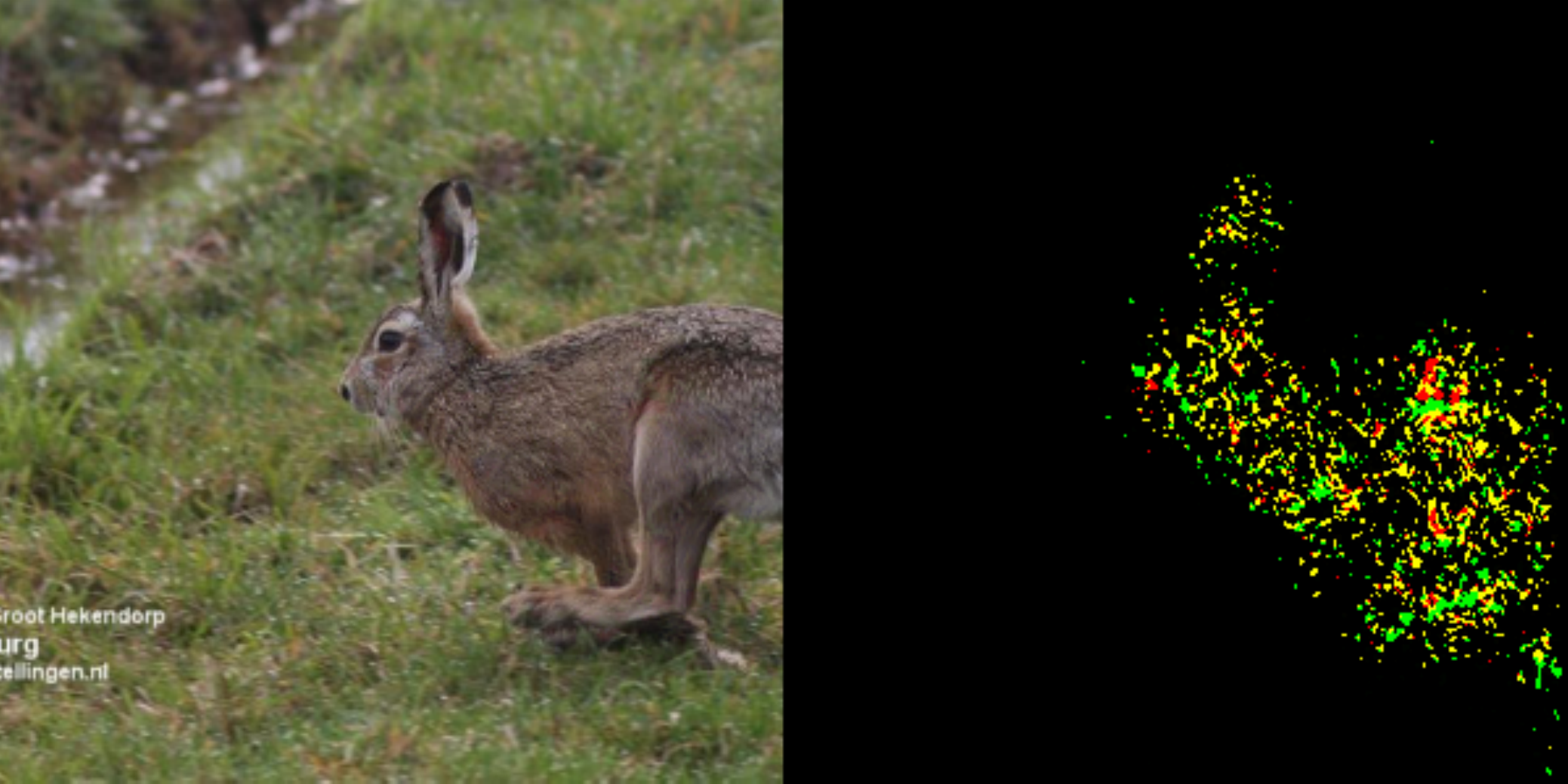}}
	\caption{Adversarial examples and corresponding perturbations constructed with our method under different perturbation magnitudes $\epsilon$. The proportions of perturbed pixels in~\ref{fig_epsilon_128} and~\ref{fig_epsilon_4} is 0.15\% and 4.36\%, respectively. The human eye can easily perceive the difference between~\ref{fig_epsilon_128} and~\ref{fig_epsilon_ori}, even though only a few pixels are perturbed.}
	\label{fig_epsilon}
\end{figure*}

\subsection{Problem analysis}
\label{section:Problem analysis}
Adversarial attacks can be divided into targeted attacks and non-targeted attacks. In this work, we focus on the targeted attack because it is more challenging. Let $f: [0,1]^{w\times h\times c}\rightarrow\mathbb{R}^{K}$ be the classification model, where $w$, $h$ and $c$ respectively denote the width, height and number of channels of the image. For an original image $\bm{x}$, $f(\bm{x})$ denotes the logits output of $K$ classes. 
The targeted adversarial attack is generally formulated as:
\begin{align}
\label{formula:adversarial attack}
\begin{split}
\min_{\bm{\delta}}\quad & \mathcal{D}(\bm{\delta})\\
\text{s.t.}\quad & \mathop{\arg\max}_{r=1,...,K}f_r(\bm{x}+\bm{\delta}) = y^{*},
\end{split}
\end{align}where $\bm{\delta} \in \mathbb{R}^{w\times h\times c}$ is the adversarial perturbation, $y^{*}$ denotes the target class, $\mathcal{D}$ is a distance function. In this paper, we focus on sparse adversarial attack using $l_0$-norm as a distance function, so the above problem encourages only a few pixels are perturbed. 

However, problem~\eqref{formula:adversarial attack} does not constrain the maximum perturbation magnitude. As shown in Figure~\ref{fig_epsilon}, when the magnitude is large (e.g. $\epsilon=128/255$), the attack can be completed by perturbing only a few pixels, but the perturbation is easy to be observed. Therefore, it is necessary to constrain the magnitudes of sparse adversarial perturbations. We formulate the sparse adversarial attack as:
\begin{align}
\label{formula:sparse adversarial attack}
\begin{split}
\min_{\bm{\delta}}\quad & \left\|\bm{\delta}\right\|_0\\
\text{s.t.}\quad & \mathop{\arg\max}_{r=1,...,K}f_r(\bm{x}+\bm{\delta}) = y^{*}\\
& \bm{\delta} \in [-\epsilon, \epsilon]^{w\times h\times c},
\end{split}
\end{align}where $\epsilon$ denotes the maximum perturbation magnitude allowed, which is the $l_{\infty}$-norm of the perturbation. Therefore, the above problem perturb as few pixels as possible within a specified maximum perturbation magnitude.

Unfortunately, since the $l_0$-norm is not differentiable, it is difficult to optimize the problems directly. 
However, we can regard sparse adversarial attacks from another perspective. 
The general training process of the classification model is to iteratively update the weights in the network to make the classification loss as small as possible. On the contrary, the network weights in the process of adversarial attacks are fixed, and the input image is iteratively updated to make the adversarial loss as small as possible. Therefore, we can look at adversarial attacks from the perspective of model training. We regard the weights in the network as fixed inputs and the input images as the weights to be updated. Thus, the problem of sparse adversarial attacks can be regarded as a problem of neural network pruning, that is, only part of the weights (pixels) are used. 
In other words, referring to the idea of neural network pruning, perturbations can also be pruned.
AutoPruner~\citep{luo2020autopruner} is a neural network automatic pruning method. It adds a branch in the usual model training process, and the branch outputs a mask to automatically select filters. With reference to this idea, we added a branch in the usual attack process, which also outputs a mask to automatically select pixels.


\subsection{Framework of AutoAdversary}

\begin{figure*}
	\centering
	\includegraphics[width=0.92\linewidth]{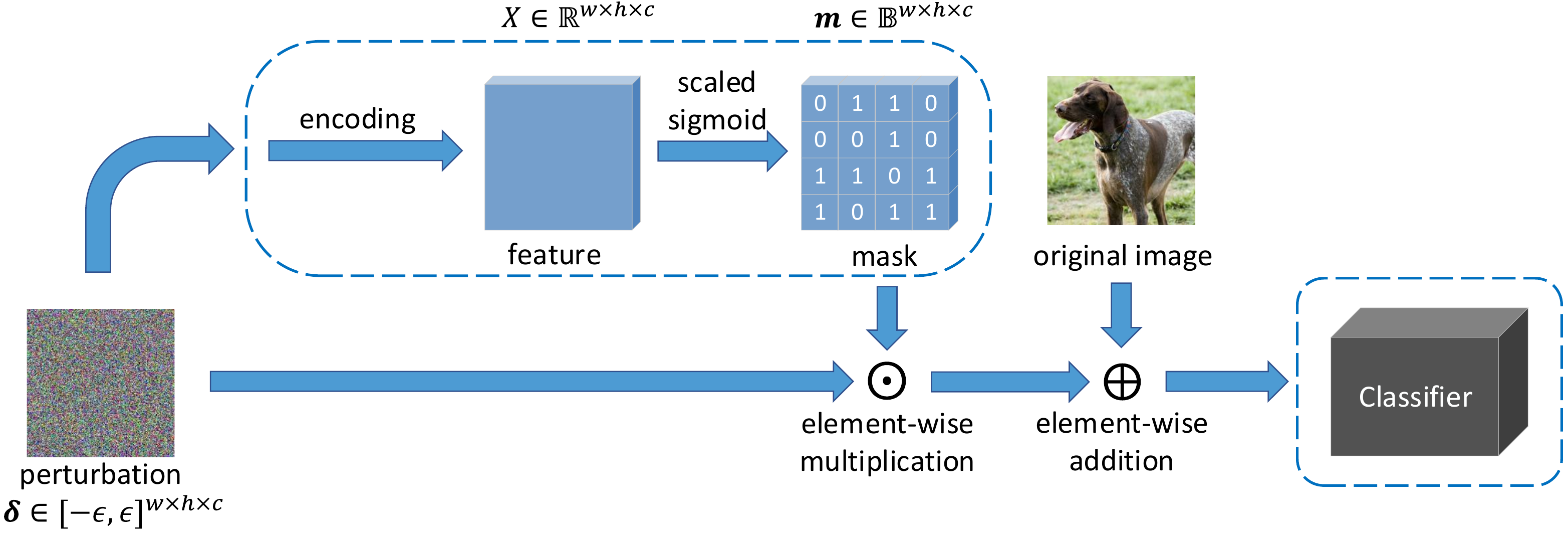}
	\caption{The framework of AutoAdversary. Given an adversarial perturbation, we encode it through a trainable neural network to generate a tensor of the same size. Then, we use the scaled sigmoid function to generate an approximately binary mask. Multiply the adversarial perturbation by the mask to generate a pruned perturbation, which is added to the original image and input into the classification model. By gradually increasing the scaling factor of the scaled sigmoid function during the training process, a binarized mask will eventually be produced, which also produces the sparse adversarial perturbation.}
	\label{fig_framework}
\end{figure*}

The framework of AutoAdversary is shown in Figure~\ref{fig_framework}. AutoAdversary can be seen as adding a branch that includes encoding and binarization to the usual adversarial attack process. The branch takes the adversarial perturbation as input and generates an approximately binary mask with the same size. Then we multiply the mask and the adversarial perturbation element by element to generate a pruned perturbation, which is added to the original image and then input into the classification model. Then, according to the sparsity and attack effect, the adversarial perturbation and the encoder are jointly optimized. By gradually forcing the scaled sigmoid function to output binary values, the perturbation of some pixels added to the original image eventually becomes 0. Thus, this method can automatically select the pixels to be perturbed during the attack.

\subsubsection{Encoding}
Let $\mathcal{H}:\mathbb{R}^{w\times h\times c} \rightarrow \mathbb{R}^{w\times h\times c}$ denote a neural network, which is used to encode the perturbation. 
$\mathcal{H}(\bm{\delta}) \in \mathbb{R}^{w\times h\times c} $ represents the tensor obtained after encoding the perturbation $\bm{\delta}$ with $\mathcal{H}$.

When the image size is small, we directly use a fully-connected layer as the encoder, and its weights are denoted as $\mathcal{W} \in \mathbb{R}^{(whc)\times(whc)}$. Specifically, the number of weights on CIFAR-10 is $9.44\times 10^{6}$. When the image size is large, using a fully-connected layer cause too many weights to train. Considering that the input and output dimensions must be the same, we use the classical image segmentation network U-net~\citep{ronneberger2015u} as the encoder, and the number of weights is $3.10\times 10^{7}$. 
This small number of weights keeps our method relatively fast on ImageNet.

It should be noted that in this paper and many other methods~\citep{fan2020sparse}, the three channels of one position in an RGB image are regarded as three independent units. In some other methods, the three units are regarded as a whole, i.e., if any unit is perturbed, the corresponding position is considered to be perturbed. By changing the output dimension of the encoder $\mathcal{H}$, we can flexibly change the channel independence to accommodate these two different assumptions. Please refer to the supplementary materials for a specific description.

\subsubsection{Binarization}
\label{section:Binarization}
The elements of the encoded tensor are real numbers, but the elements of the mask we need are binary. To keep continuity and differentiability, we use the scaled sigmoid function to generate an approximate binary mask:
\begin{align}
\label{formula:sigmoid}
\bm{m} = \text{sigmoid}(\alpha\cdot \mathcal{H}(\bm{\delta})),
\end{align}where $\bm{m} \in [0,1]^{w\times h \times c}$ denotes the approximate binary mask and $\alpha$ is the scaling factor which controls the degree of binarization. 

When $\alpha$ is too small,~\eqref{formula:sigmoid} is not enough to binarize the elements of mask. 
When $\alpha$ is too large, the elements of mask can be binarized, but this results in the selection of pixels has been determined before the training, which makes our method degenerate into randomly selecting pixels.

Therefore, like AutoPruner~\citep{luo2020autopruner}, we gradually increase $\alpha$ from $\alpha_{\text{start}}$ to $\alpha_{\text{end}}$ during the training process to ensure that the elements of mask can converge to binary values and avoid the method degenerating into random pixels selection.

\subsubsection{Loss function}
According to the foregoing, our goal is to make the mask as sparse as possible, and to make the pruned perturbations fool the target model as much as possible. The most common sparse method is the convex relaxation $l_1$-norm, which is defined by $\left\|\bm{m}\right\|_1$. Therefore, we can formulate our loss function as:
\begin{align}
\label{formula:loss}
\mathcal{L} = \mathcal{L}_{\text{adv}}(f(\bm{x}+\bm{\delta} \odot \bm{m}), y^{*}) + \lambda \frac{\left\|\bm{m}\right\|_1}{N},
\end{align}where $\odot$ denotes the element-wise multiplication, $\mathcal{L}_{\text{adv}}(\cdot, \cdot)$ is the adversarial loss, usually the cross-entropy loss. Since the elements of $\bm{m}$ are approximately binary, the second term approximately represents the sparsity of the mask, where $N=w\times h\times c$ denotes the size of the mask. $\lambda$ is a dynamic parameter that balances these two terms, which is calculated by the following formula:
\begin{align}
\label{formula:lambda}
\lambda = C + \frac{\gamma}{N} \sum_{i=1}^{N}\mathbb{I}(m_i > 0.5),
\end{align}
where $C>0$ is a hyperparameter, which is the minimum value of $\lambda$, $m_i$ is the $i$-th element of the mask $\bm{m}$, $\gamma > 0$ is a hyperparameter. So the second term approximately represents the sparsity of the current mask. If the current mask is not sparse enough, $\lambda$ is relatively large, so AutoAdversary could pay more attention to improving the sparsity of the mask. When the mask is sparse enough, $\lambda$ is relatively small, so AutoAdversary could focus more on the adversarial attack.

In this paper, we try to make the perturbation as sparse as possible. However, the degree of sparsity can be specified by slightly modifying~\eqref{formula:loss}. Details are in the supplementary materials.

\subsubsection{Update perturbation and encoder}
Since we use the mask to make the final perturbation sparse, we do not need to consider its own sparsity. Therefore, we can use dense attack methods to quickly update $\bm{\delta}$. Moreover, considering the constrain of $\bm{\delta}$ in problem~\eqref{formula:sparse adversarial attack}, we combine PGD~\citep{madry2018towards} and MI-FGSM~\citep{dong2018boosting} because they attack under $l_{\infty}$-norm constrain. Therefore, we can constrain both the $l_0$-norm and the $l_{\infty}$-norm of the perturbation. Specifically, the update formula of $\bm{\delta}$ is as follows:
\begin{align}
\label{formula:attack_grad}
& \bm{g}_{t+1} = \mu\cdot\bm{g}_t + \frac{\nabla_{\bm{\delta}}\mathcal{L}}{\left\| \nabla_{\bm{\delta}}\mathcal{L} \right\|_1}\\
\label{formula:attack}
& \bm{\delta}_{t+1} = \text{Clip}_{\bm{\epsilon}}\{\bm{\delta}_{t} - \beta\cdot\text{sign}(\bm{g}_{t+1})\},
\end{align}where $\mathcal{L}$ is the loss defined by~\eqref{formula:loss}, $\mu$ denotes the momentum decay factor, $\beta$ represents the update step, and $\text{Clip}_{\epsilon}\{\cdot\}$ is used to project adversarial perturbation into the $l_{\infty}$-ball of radius $\epsilon$. Just like PGD~\citep{madry2018towards}, the initial $\bm{\delta}$ is a uniformly random point in the $l_{\infty}$-ball.

As for the update of the encoder $\mathcal{H}$, we directly use stochastic gradient descent (SGD) with momentum. To summarize, the perturbation is trained simultaneously with the encoder. The completion of training means the completion of the attack. The specific process is shown in Algorithm~\ref{alg:algorithm}.

\begin{algorithm}[tb]
\caption{AutoAdversary}
\label{alg:algorithm}
\textbf{Input}: Original image $\bm{x}$, target class $y^{*}$, target model $f$\\
\textbf{Parameter}: Max iterations $T$, threshold $\epsilon$, balance parameters $\gamma$ and $C$, scaling factors $\alpha_{\text{start}}$ and $\alpha_{\text{end}}$, decay factor $\mu$, update step $\beta$\\
\textbf{Output}: Sparse adversarial example $\bm{x}^{\text{adv}}$
\begin{algorithmic}[1] 
\STATE Sample $\bm{\delta}_{0}$ uniformly from the $l_{\infty}$-ball of radius $\epsilon$;
\STATE Randomly initializes the encoder $\mathcal{H}_{0}$;
\STATE $\alpha_{0} \leftarrow \alpha_{\text{start}}$;
\STATE $\bm{g}_{0} \leftarrow \bm{0}$;
\FOR{$t = 0,1,..., T-1$}
\STATE $\bm{m}_{t} \leftarrow \text{sigmoid}(\alpha_{t} \cdot \mathcal{H}_{t}(\bm{\delta}_{t}))$;
\STATE Calculate the dynamic parameter $\lambda_{t}$ based on $\bm{m}_{t}$ and Eq.~\eqref{formula:lambda};
\STATE Calculate the loss $\mathcal{L}$ based on $\bm{m}_{t}$, $\bm{\delta}_{t}$, $\lambda_{t}$ and Eq.~\eqref{formula:loss};
\STATE Update $\bm{g}_{t+1}$ by Eq.~\eqref{formula:attack_grad};
\STATE Update $\bm{\delta}_{t+1}$ by Eq.~\eqref{formula:attack};
\STATE Update the encoder $\mathcal{H}_{t+1}$ based on the loss $\mathcal{L}$ and SGD with momentum;
\STATE Update the scaling factor $\alpha_{t+1}$ using the algorithm in~\citep{luo2020autopruner};
\ENDFOR
\STATE $\bm{m}_{T} \leftarrow \text{sigmoid}(\alpha_{T} \cdot \mathcal{H}_{T}(\bm{\delta}_{T}))$;
\STATE $\bm{x}^{\text{adv}} \leftarrow \bm{x} + \bm{\delta}_{T} \odot \bm{m}_{T}$;
\STATE \textbf{return} $\bm{x}^{\text{adv}}$.
\end{algorithmic}
\end{algorithm}

\section{Experimental results}
\label{section:Experimental results}
We conduct experiments on CIFAR-10~\citep{krizhevsky2009learning} and ImageNet~\citep{deng2009imagenet}. We compared our method with several state-of-the-art sparse adversarial attack algorithms, including four two-stage algorithms (JSMA~\citep{papernot2016limitations}, C$\&$W-$l_0$~\citep{carlini2017towards}, PGD-$l_0+l_{\infty}$~\citep{croce2019sparse}, GreedyFool~\citep{dong2020greedyfool}), and one one-stage algorithm similar to ours (SAPF~\citep{fan2020sparse}).

\subsection{Experimental settings}
\subsubsection{Classification models and datasets}
For the target classifier, on CIFAR-10, we train a VGG19 model~\citep{simonyan2014very} with an input size of $32\times 32\times 3$. The model achieves $79\%$ top-1 classification accuracy on the validation set. On ImageNet, we use a pre-trained Inception-v3 model~\citep{szegedy2016rethinking} with $78\%$ top-1 classification accuracy. The input size of the model is $299 \times 299 \times 3$. We select the images correctly classified by their corresponding target models from the first 1000 images in CIFAR-10 validation set and the random 100 images in ImageNet validation set, then construct the adversarial examples from these images. The target label of each image is randomly selected and is not equal to the true label.

\subsubsection{Implementation details}
\label{section:Implementation details}



There are some important implementation details to note. As mentioned in Section~\ref{section:Binarization}, we gradually increase the scaling factor $\alpha$ from $\alpha_{\text{start}}$ to $\alpha_{\text{end}}$ in the optimization process to make the mask binary. 
Referring to the strategy of AutoPruner~\cite{luo2020autopruner}, we first find $\alpha_{\text{end}}$ which can produce a true binary output in the scaled sigmoid function. This step can be done quickly, because we only need to try several numbers on two or three images to determine a good enough $\alpha_{\text{end}}$. Specifically, $\alpha_{\text{end}}$ is set to $100$ on CIFAR-10 and $10$ on ImageNet. The change of $\alpha_{\text{start}}$ has little effect on the results, so we just set $\alpha_{\text{start}}$ to $0.1$ on both CIFAR10 and ImageNet. 
Due to the space limit, other implementation details are presented in supplementary materials.

\subsubsection{Evaluation metrics}
The average $l_p$-norm ($p = 0, 2, \infty$) of perturbations, attack success rate (ASR), and average time spent of each image are used to evaluate the performance of different methods. Due to the limitation of space, the less commonly used $l_1$-norm is only compared in the supplementary material.

As mentioned in Section~\ref{section:Problem analysis}, the difficulty of sparse attacks is different under different maximum perturbation magnitudes. However, some methods can only control the sparsity (i.e., $l_0$-norm of the perturbation), but cannot restrict the maximum perturbation magnitude (i.e., $l_{\infty}$-norm of the perturbation). On the contrary, some other methods can only control $l_{\infty}$-norm but cannot specify $l_0$-norm. For a fair comparison, we first run our proposed AutoAdversary under a specified perturbation magnitude $\epsilon$, then run other methods with reference to $\epsilon$ or $l_0$-norm of our perturbations. Specifically, for those methods that can only specify $l_{\infty}$-norm, we set $l_{\infty} = \epsilon$, then compare the $l_0$-norm, $l_2$-norm and ASR. For those methods that can only specify $l_0$-norm, we let them run with reference to $l_0$-norm of our perturbations, then compare the $l_2$-norm, $l_{\infty}$-norm and ASR.

\subsection{Experimental comparisons}
\subsubsection{Results on CIFAR-10}
The average $l_p$-norm of perturbations and the ASR on CIFAR-10 under two values of $\epsilon$ are given in Table~\ref{tab:comparisons_cifar}. When $\epsilon = 8/255$, our proposed AutoAdversary achieves $100\%$ ASR and the $l_0$-norm is only $320.1$ (10.42\% pixels). When $\epsilon = 16/255$, our method achieves $100\%$ ASR and the $l_0$-norm is only $131.3$ ($4.27\%$ pixels). JSMA and GreedyFool can directly specify the $l_{\infty}$-norm, so we make these two methods run under the condition that $l_{\infty}=\epsilon$. We can see that the ASR of these two methods is not higher than our method, and the $l_0$-norm, and $l_2$-norm are obviously larger than our method. PGD-$l_0+l_{\infty}$ needs to specify both $l_{\infty}$-norm and $l_0$-norm, so we make the method run under the condition that $l_{\infty}=\epsilon$, and try to find a suitable $l_0$-norm. However, even a large $l_0$-norm dose not make the method achieve 100\% ASR. C$\&$W-$l_0$ and SAPF can only specify $l_0$-norm, so we run these two methods with reference to our $l_0$-norm. It can be seen that the $l_0$-norm of these two methods is almost the same as ours. But the $l_{\infty}$-norm of our method are significantly smaller than these two methods. In other words, the maximum perturbation magnitude generated by our method are smaller and less easily be perceived, that is, C$\&$W-$l_0$ and SAPF obtain sparsity similar to our method under a more relaxed condition. These results all demonstrate the superiority of our proposed method.
\begin{table}
\caption{Results of targeted sparse adversarial attack on CIFAR-10.}
\label{tab:comparisons_cifar}
\centering
\begin{tabular}{lrrrrrrrr}
\toprule
Method 						& \multicolumn{4}{c}{$\epsilon=8/255$} 											& \multicolumn{4}{c}{$\epsilon=16/255$} \\
\cmidrule(r){2-5}
\cmidrule(r){6-9}
                        & ASR (\%)   		& $l_0$  			& $l_2$  			& $l_{\infty}$	 	& ASR (\%)  		& $l_0$ 			& $l_2$  			& $l_{\infty}$  	\\
\midrule
JSMA                    & 78.9    			& 440.8 			& 0.611 		 	& 0.031         	& 97.3    			& 247.7 			& 0.896 			& 0.063     		\\
PGD-$l_0+l_{\infty}$    & 73.9    			& 1199.7			& 1.078  			& 0.031     	   	& 72.8    			& 498.0 			& 1.390 			& 0.063   			\\
GreedyFool              & 100.0   			& 486.2 			& 0.547 		 	& 0.031         	& 100.0   			& 238.3 			& 0.707 			& 0.063   			\\
C$\&$W-$l_0$            & 100.0   			& 326.6 			& 0.542  			& 0.068     	   	& 100.0   			& 136.7 			& 0.691 			& 0.118   			\\
SAPF                    & 100.0  			& 321.8 			& \textbf{0.523}	& 0.085          	& 100.0 			& 133.7 			& 0.718 			& 0.159     		\\
\textbf{AutoAdversary}  & \textbf{100.0} 	& \textbf{320.1} 	& 0.532  			& \textbf{0.031} 	& \textbf{100.0}  	& \textbf{131.3}	& \textbf{0.689} 	& \textbf{0.063}	\\
\bottomrule
\end{tabular}
\end{table}

\subsubsection{Results on ImageNet}
Since JSMA has high computational complexity on large-scale images, we do not compare it. The results of different methods on ImageNet are given in Table~\ref{tab:comparisons_imagenet}. When $\epsilon = 8/255$, our proposed AutoAdversary achieves $100\%$ ASR and the $l_0$-norm is only $5591.5$ (2.08\% pixels). When $\epsilon = 4/255$, our method achieves $100\%$ ASR and the $l_0$-norm is only $12855.0$ (4.79\% pixels). It can be seen that PGD-$l_0+l_{\infty}$ and GreedyFool maintain the same $l_{\infty}$-norm as our method, but the ASR of PGD-$l_0+l_{\infty}$ is far less than 100\%, and the $l_0$-norm, and $l_2$-norm of these two methods are significantly larger than our method. 
The official code provided by SAPF cannot converge on ImageNet, so we use the results provided in their paper directly. Although the $l_2$-norm of SAPF is very small, its $l_0$-norm and $l_{\infty}$-norm are much larger than ours, that is to say, the performance of SAPF in sparsity is not as good as ours. We run C$\&$W-$l_0$ with reference to our $l_0$-norm, so this method has almost the same $l_0$-norm as our method, and reach 100\% ASR. But the $l_{\infty}$-norm of our method is much smaller than that of C$\&$W-$l_0$ in both $\epsilon$ values. In other words, the maximum perturbation magnitude of C$\&$W-$l_0$ is larger than ours, that is, C$\&$W-$l_0$ achieves a sparsity similar to our method under a more relaxed condition. These results all demonstrate the superiority of our proposed method.
\begin{table}
\caption{Results of targeted sparse adversarial attack on ImageNet.}
\label{tab:comparisons_imagenet}
\centering
\begin{threeparttable}
\begin{tabular}{lrrrrrrrr}
\toprule
Method 					& \multicolumn{4}{c}{$\epsilon = 8/255$} 										& \multicolumn{4}{c}{$\epsilon = 4/255$} \\
\cmidrule(r){2-5}
\cmidrule(r){6-9}
                        & ASR (\%) 			& $l_0$ 			& $l_2$  			& $l_{\infty}$ 		& ASR (\%)   		& $l_0$ 			& $l_2$  			& $l_{\infty}$  	\\
\midrule
PGD-$l_0+l_{\infty}$    & 50.0    			& 11989.0 			& 2.957				& 0.031      	  	& 62.2 				& 22980.9 			& 2.116 			& 0.016    			\\
GreedyFool              & 100.0   			& 12454.5 			& 2.663				& 0.031     	  	& 100.0				& 26107.7 			& 2.083				& 0.016    			\\
SAPF\tnote{*}			& 100.0 			& 37275.0 			& \textbf{0.586} 	& 0.038      	  	& 100.0 			& 37275.0 			& \textbf{0.586} 	& 0.038    				\\
C$\&$W-$l_0$            & 100.0   			& 5697.4  			& 1.257				& 0.068           	& 100.0 			& 13028.2 			& 1.084 			& 0.051     		\\
\textbf{AutoAdversary}  & \textbf{100.0} 	& \textbf{5591.5}   & 2.223  			& \textbf{0.031}	& \textbf{100.0}	& \textbf{12855.0}	& 1.694 			& \textbf{0.016} 	\\
\bottomrule
\end{tabular}
\begin{tablenotes}
\footnotesize
\item[*] The official code of SAPF~\citep{fan2020sparse} cannot converge on ImageNet, so we use the results in their paper directly. (https://github.com/wubaoyuan/Sparse-Adversarial-Attack)
\end{tablenotes}
\end{threeparttable}
\end{table}

\subsubsection{Speed comparison}
Here we compare the average time spent on each image by the different methods in Table~\ref{tab:speed}.
All methods run on a NVIDIA RTX 2080Ti GPU.
Since PGD-$l_0+l_{\infty}$ can attack multiple images in one batch, its average time is the shortest, but its ASR is extremely low. SAPF is similar to our method, but it is already very slow when attacking small-size images because it requires a lot of iterations. PGD-$l_0+l_{\infty}$, GreedyFool and C$\&$W-$l_0$ both contain a heuristic selection process, so as the image size increases, the time they spend increases dramatically. The attack time required for our proposed AutoAdversary has only increased by 771.40\% from CIFAR-10 to ImageNet. In short, the speed of our method is competitive on both CIFAR-10 and ImageNet, while guaranteeing better sparsity. Moreover, the speed of our method dose not slow down sharply with the increase of image size.
\begin{table}
\caption{Average time spent attacking each image on CIFAR-10 and ImageNet when $\epsilon = 8/255$. The last column represents the growth rate of time spent from CIFAR-10 to ImageNet.}
\label{tab:speed}
\centering
\begin{tabular}{lrrr}
\toprule
Method                  & CIFAR-10 & ImageNet & Growth rate (\%) \\
\midrule
SAPF                    & 395.87 & -        & -      \\
JSMA                    & 151.16 & -        & -      \\
C$\&$W-$l_0$            & 33.47  & 1102.61 & 3194.32 \\
GreedyFool              & 2.24   & 99.40   & 4337.50 \\
PGD-$l_0+l_{\infty}$    & \textbf{0.49} & \textbf{15.68} & 3100.00 \\
\textbf{AutoAdversary}  & 14.58  & 127.05 & \textbf{771.40}  \\
\bottomrule
\end{tabular}
\end{table}

\subsection{Ablation study}

\subsubsection{Component analysis of AutoAdversary}


We further analyze the contributions of each part of our method. In the following, we use "Dense" to denote the dense attack method we use to update $\bm{\delta}$, "Random" to denote random pixel selection, "$l_1$-$\bm{\delta}$" to denote directly constraint the $l_1$-norm of the perturbation, "$l_1$-$\bm{m}$" to denote constraint the $l_1$-norm of the mask, "Encoder" to denote encoding perturbations with a neural network, and "Binarization" to denote binarization with the scaled sigmoid function.

From the results shown in Table~\ref{tab:mask ablation}, we observe that "Dense" can reach 100\% ASR, that is, the dense attack method we use to update perturbations has a good attack effect. "Dense + Random" means to select some pixels randomly, and then do a dense attack on these pixels. The ASR of this random method is only 23.38\%, which indicates that our perturbed pixels are not random. As shown in equation~\eqref{formula:loss}, we add the $l_1$-norm of the mask to the adversarial loss function to make the mask sparse. As a contrast, instead of generating the mask, we try to generate the sparse perturbation directly by adding the $l_1$-norm of the perturbation to the adversarial loss function, which is denoted as "Dense + $l_1$-$\bm{\delta}$". This method cannot achieve 100\% ASR, and its $l_0$-norm is 2435.471 (79.28\% pixels). Therefore, only replacing the minimization of $l_0$-norm with the minimization of $l_1$-norm cannot solve the problem of sparse adversarial attacks, which also indicates that the added branch including encoding and binarization in our method is effective. Furthermore, to verify the importance of the encoder, we directly input the perturbation into our binarization module to generate the mask without any encoder, and this method is denoted as "Dense + $l_1$-$\bm{m}$ + Binarization". Without encoders, sparse perturbations can be generated, but the ASR is low. This is because our binarization is a sigmoid function, which makes the perturbation cannot be negative, resulting in the solution space being reduced. Therefore, using an encoder to encode the perturbation first prevents reducing the solution space. Finally, after adding the encoder, the ASR and sparsity are the best, which indicates the importance of the encoder.

\begin{table}
\caption{The contribution of each part in AutoAdversary. The ASR and average $l_p$-norm here are calculated on CIFAR-10 ($\epsilon=8/255$).}
\label{tab:mask ablation}
\centering
\begin{tabular}{lrrrr}
\toprule
Method            								& ASR (\%)   		& $l_0$ 			& $l_2$ 			& $l_{\infty}$ \\
\midrule
Dense                  							& 100.0 			& 3032.3 		 	& 1.628 			& 0.031 			\\
Dense + Random    		   						& 23.4 				& 982.6 			& 0.948 			& 0.031  			\\
Dense + $l_1$-$\bm{\delta}$         			& 96.2 				& 2435.5	 		& 0.785 			& 0.031  			\\
Dense + $l_1$-$\bm{m}$ + Binarization         	& 91.8			 	& 462.9 			& 0.638 			& 0.031  			\\
Dense + $l_1$-$\bm{m}$ + Binarization + Encoder & \textbf{100.0} 	& \textbf{320.1}	& \textbf{0.532} 	& \textbf{0.031}   	\\
\bottomrule
\end{tabular}
\end{table}

\subsubsection{Selection of encoder structure}
We further analyze the influence of different encoder structures in Table~\ref{tab:encoder}. 
The only requirement for the encoder is that the input and output are of the same size, so we compare the simplest fully-connected network and two classic image segmentation networks.
In the following, "U-net", "FCN" and "FC" respectively indicate that U-net~\citep{ronneberger2015u}, FCN~\citep{long2015fully} and fully-connected layer are used as encoders. Interestingly, "U-net", "FCN" and "FC" all achieve similar good results. This demonstrates that our method does not depend on a specific structure of encoder, so it can be used as a general framework for sparse adversarial attacks.


\begin{table}
\caption{ASR and average $l_p$-norm of perturbations on CIFAR-10 ($\epsilon=8/255$) using encoders with different structures.}
\label{tab:encoder}
\centering
\begin{tabular}{lrrrrr}
\toprule
Encoder            	& ASR (\%)	& $l_0$ & $l_2$  & $l_{\infty}$ \\
\midrule
U-net         		& 100.0 	& 343.1 & 0.549 & 0.031  		\\
FCN         		& 100.0 	& 345.7 & 0.551 & 0.031  		\\
FC         			& 100.0 	& 320.1 & 0.532 & 0.031   		\\
\bottomrule
\end{tabular}
\end{table}

\section{Conclusions}

In this paper, we propose a novel end-to-end sparse adversarial attack method. Our method jumps out of the fixed mode, and no longer relies on the hand-crafted pixel importance evaluation criteria, but automatically selects pixels for good sparsity and attack effect. Experimental results demonstrate the superiority of our proposed method to state-of-the-art sparse adversarial attack methods. Moreover, since there is no heuristic process of pixel selection, the speed of our method does not slow down excessively due to the increase of image size. Further study reveals that our method does not depend on a specific encoder structure, so it can be used as a general framework for sparse adversarial attack.



\bibliographystyle{plainnat}
\bibliography{neurips_2021}



\end{document}